\documentclass[journal]{IEEEtran}

\usepackage{amsmath,graphicx}
\usepackage{amssymb}
\usepackage{cases}
\usepackage{cite}
\usepackage{amsthm}
\usepackage{color}
\usepackage{lineno,hyperref}
\usepackage{booktabs}

\author{Fanghui Liu, Tao Zhou, Irene Y.H. Gu and Jie Yang\footnote{Corresponding author: jieyang@sjtu.edu.cn }}
\title{Visual Tracking via Nonnegative Regularization Multiple Locality Coding}
\begin{document}
\maketitle
\begin{abstract} 
\noindent This paper presents a novel object tracking method based on approximated Locality-constrained Linear Coding (LLC).
  Rather than using a non-negativity constraint on encoding coefficients to guarantee these elements nonnegative, in this paper, the non-negativity constraint is substituted for a conventional $\ell_2$ norm regularization term in approximated LLC to obtain the similar nonnegative effect.
  And we provide a detailed and adequate explanation in theoretical analysis to clarify the rationality of this replacement.
  Instead of specifying fixed K nearest neighbors to construct the local dictionary, a series of different dictionaries with pre-defined numbers of nearest neighbors are selected.
  Weights of these various dictionaries are also learned from approximated LLC in the similar framework.
  In order to alleviate tracking drifts, we propose a simple and efficient occlusion detection method.
  The occlusion detection criterion mainly depends on whether negative templates are selected to represent the severe occluded target.
  Both qualitative and quantitative evaluations on several challenging sequences show that the proposed tracking algorithm achieves favorable performance compared with other state-of-the-art methods.

\end{abstract}


\section{Introduction}
\vspace{-0.25cm}
\label{sec:intro}

Visual tracking is an indispensable part of computer vision with wide ranging applications, such as video surveillance,
vehicle navigation and medical imaging \cite{henriques2015high,Wu2015}.
While much effort \cite{Adam2006,Gong2014,Zhou2015,Wang2013} has been made on object tracking in the last few years, it seems difficult to find lasting solutions to enduring problems due to intrinsic factors (e.g. shape deformation and pose variation) and extrinsic factors (e.g. occlusions and varying illumination).

Recently, sparse representation \cite{wright2009robust,Zhang2013} has been successfully applied in visual tracking, early stemming from the $\ell_1$ tracker proposed by Mei $et~al.$ \cite{Mei2011} and their improved version \cite{mei2011minimum}.
The $\ell_1$ method represents a target by a sparse linear combination of the target templates and trivial templates, and then solves it using a $\ell_1$-regularized least squares method.
The accelerated proximal gradient approach \cite{Bao2012} is used to solve $\ell_1$ norm minimization efficiently.
However, because of adopting the global sparse appearance model, these methods are less effective in handling heavy occlusions.
Different from them, a local sparse appearance model \cite{Jia2012,wang2012online} is introduced to enhance target representation and tracking robustness.
Local patches inside a possible target candidate are sparsely represented with local patches in the dictionary templates.
Joint sparse appearance model \cite{Hong2013,zhang2012low,Zhang2014c} exploits the intrinsic relationship among particles to represent the target jointly.

Locality-constrained Linear Coding (LLC) \cite{Wang2010} is proposed to represent local appearance in tracking framework \cite{Liu2013} because of its excellent performance (eg. similarity in feature space and close-form solution).
It utilizes the sparse histograms of sparse coefficients and local optimal search scheme for object tracking.
However, this method  just uses a static local dictionary and does not provide additional constraint on encoding coefficients, leading to tracking drift easily.
The spatial layout information is embedded in coding stage on appearance model \cite{Liu2014}.

Motivated by the previous work, we aim to develop a more robust approximated LLC tracker, especially when the non-negativity constraint on encoding coefficients is taken into consideration.
The main contributions of this paper are as follows.
(1) A $\ell_2$ norm regularization term is introduced into approximated LLC, instead of the non-negativity constraint, to guarantee encoding coefficients nonnegative by choosing a regularization parameter.
(2) Rather than using a static local sparse dictionary, a series of local dictionaries with pre-defined different numbers of nearest neighbors are provided.
Weights of a linear combination of these dictionaries are learned from approximated LLC, and that is similar with solving encoding coefficients in the same framework.
(3) To mitigate drifting problem, the occlusion detection criterion depends on whether negative templates are used to represent the target.
Once negative templates are selected to reconstruct the target, we conclude that the target suffers from severe occlusion.
Experiments on some public sequences compared with several prevalent tracking methods show the effectiveness and robustness of the proposed tracking algorithm.

The remainder of the paper is organized as follows. Section \ref{sec:related} gives the relevant related work of the proposed method. Section \ref{sec:proposed} gives the details of the proposed method. Section \ref{sec:experiment} shows the experimental results from the proposed method, with comparisons to eight existing state-of-the-art methods. Finally, conclusion is given in Section \ref{sec:conclusion}.
\vspace{-0.25cm}

\section{Preliminaries}
\label{sec:related}
\subsection{Particle Filter in Tracking Framework}
The implicit rationale behind particle filter \cite{Kwon2009,Zhong2012a} is to estimate the posterior distribution $p({\rm \textbf{x}_t}|\textbf{z}_{1:t})$ approximately by a finite set of random sampling particles.
Given some observed image patches at $t$-th frame ${\textbf{z}}_{1:t}=\{\textbf{z}_1,\textbf{z}_2,...,\textbf{z}_{t}\}$, the state of the target $\textbf{x}_t$ can be estimated recursively.
\begin{small}
\begin{equation}
p(\textbf{x}_t|\textbf{z}_{1:t})\propto p(\textbf{z}_t|\textbf{x}_t)\int p(\textbf{x}_t|\textbf{x}_{t-1})p(\textbf{x}_{t-1}|\textbf{z}_{1:t-1}) \mathrm{d}\textbf{x}_{t-1}
\end{equation}\vspace{0.1cm}
\end{small}
Let $\textbf{x}_t=[l_x,l_y,\theta,s,\alpha,\phi]^T$, where $l_x,l_y,\theta,s,\alpha,\phi$ denote translations in the direction of $x$,$y$, rotation angle, scale, aspect ratio, and skew respectively.
$p(\textbf{x}_t|\textbf{x}_{t-1})\sim\mathcal{N}(\mu, \sigma^2)$, referred to as the motion model, denotes state transition between two consecutive frames.
The observation model $p(\textbf{z}_t|\textbf{x}_t)$ reflects the similarity between a target candidate and the target templates.
Thus, the optimal state at $t$-th frame is obtained by maximizing the observation model:
 \begin{equation}
\textbf{x}^*_t = \mathop{\mathrm{argmax}}\limits_{x}  p(\textbf{z}_t|\textbf{x}_t)
\end{equation}
In our method, the observation model is formulated from the reconstruction error by the multiple local dictionaries using approximated LLC.

\subsection{Locality-constrained Linear Coding}
LLC applies locality constraint to select similar basis of local image descriptors from a codebook, and learns a linear combination weight of these basis $\textbf{B}$ to reconstruct each descriptor $\textbf{x}_i$.\vspace{-0.3cm}
 \begin{equation}\label{llc}
 \begin{split}
&\mathop{\mathrm{min}}\limits_{{\rm \textbf{C}}} \sum_{i=1}^{N}\|\textbf{x}_i-\textbf{B}_i\textbf{c}_i\|^2 + \lambda \|\textbf{d}_i\odot \textbf{c}_i \|^2 \\
&s.t. ~~\textbf{1}^ \mathrm{T}c_i=1,\forall i
\end{split}
\end{equation}
where $\odot$ denotes the element-wise multiplication, and $\textbf{d}_i={\rm exp}(\frac{{\rm dist}(\textbf{x}_i,\textbf{B})}{\sigma})$.
${\rm dist}(\textbf{x}_i,\textbf{B})$ represents the Euclidean distance between $\textbf{x}_i$ and $\textbf{B}$.
$\sigma$ is a scale factor parameter that controls the weight decay speed.

Approximated LLC method \cite{Wang2010}, as the simplified edition of LLC, takes local sparsity into consideration.
Instead of using the metric measure ${\rm dist}(\textbf{x}_i,\textbf{B})$, we can simply select $K$ nearest neighbors of $\textbf{x}_i$ as the local dictionary $\textbf{B}_i$.
In visual tracking process \cite{Liu2013,Liu2014}, the candidate $\textbf{y}_i$ is represented by the linear combination of several local basis vectors as the dictionary $\textbf{B}_i$ with encoding coefficients $\textbf{c}_i$ sparsely.\vspace{-0.2cm}
\begin{equation}
\begin{split}
&\mathop{\mathrm{min}}\limits_{{\rm \textbf{c}}_i} \|\textbf{y}_i-\textbf{B}_i\textbf{c}_i\|^2
~~s.t.~~ \textbf{1}^ \mathrm{T}\textbf{c}_i=1,\forall i
\end{split}
\label{allc}\vspace{-0.25cm}
\end{equation}
where $\textbf{1}$ denotes a vector with all ones and the constraint $\textbf{1}^ \mathrm{T}\textbf{c}_i=1$ ensures shift-invariant.
The closed-form solution of (\ref{allc}) is $\textbf{c}_i = [(\textbf{B}^\mathrm{T}_i -  \textbf{1}\textbf{y}^\mathrm{T}_i)(\textbf{B}_i-\textbf{y}_i\textbf{1}^\mathrm{T})]^{-1}\textbf{1}$.

\section{Proposed visual tracking algorithm}
\label{sec:proposed}
Some observed vectorization image patches($\in R^M$) at $t$-th frame ${\rm \textbf{Y}}_{1:N}=\{\textbf{y}_1,\textbf{y}_2,...,\textbf{y}_{N}\}\in R^{M\times N}$ are sampled based on particle filter framework.
In order to better capitalize on the distinction between the foreground and the background to locate the target, plenty of negative templates are collected.
The positive and negative template sets are defined as $\textbf{T}^{pos}=[\textbf{T}_1,\textbf{T}_2,...,\textbf{T}_p]$ and $\textbf{T}^{neg}=[\textbf{T}_{p+1},\textbf{T}_{p+2},...,\textbf{T}_{p+n}]$,
where $p$ and $n$ denote the number of positive and negative template sets respectively.
Generally, the tracking result in the first frame is manually chosen as a rectangle box.
Define that $\mathcal{I}(x,y)$ is the center of the rectangle box, and the initial
positive templates are sampled from an inner circular area that satisfies $\| \mathcal{I}_i-\mathcal{I}(x,y) \| < r$, where $\mathcal{I}_i$ is the center of the $i$-th sampled patch.
Similarity, negative templates are sampled from the annular region $r< \|\mathcal{I}_j-\mathcal{I}(x,y) \| < s$, where $\mathcal{I}_j$ is the center of the $j$-th sampled image, $r$ and $s$ are the inner and outer radius of the annular region respectively.
At the beginning of our tracking process, the number of positive templates and negative templates are set to $p=50$, $n=150$ respectively.
\subsection{Modification on approximated LLC}
\subsubsection{Non-negativity constraint on approximated LLC}
It might be tempting to agree that approximated LLC guarantees local sparsity and good reconstruction in visual tracking algorithm.
However, (\ref{allc}) overlooks some potential information on coefficients.
\begin{figure}
\begin{center}
\includegraphics[width=0.46\textwidth]{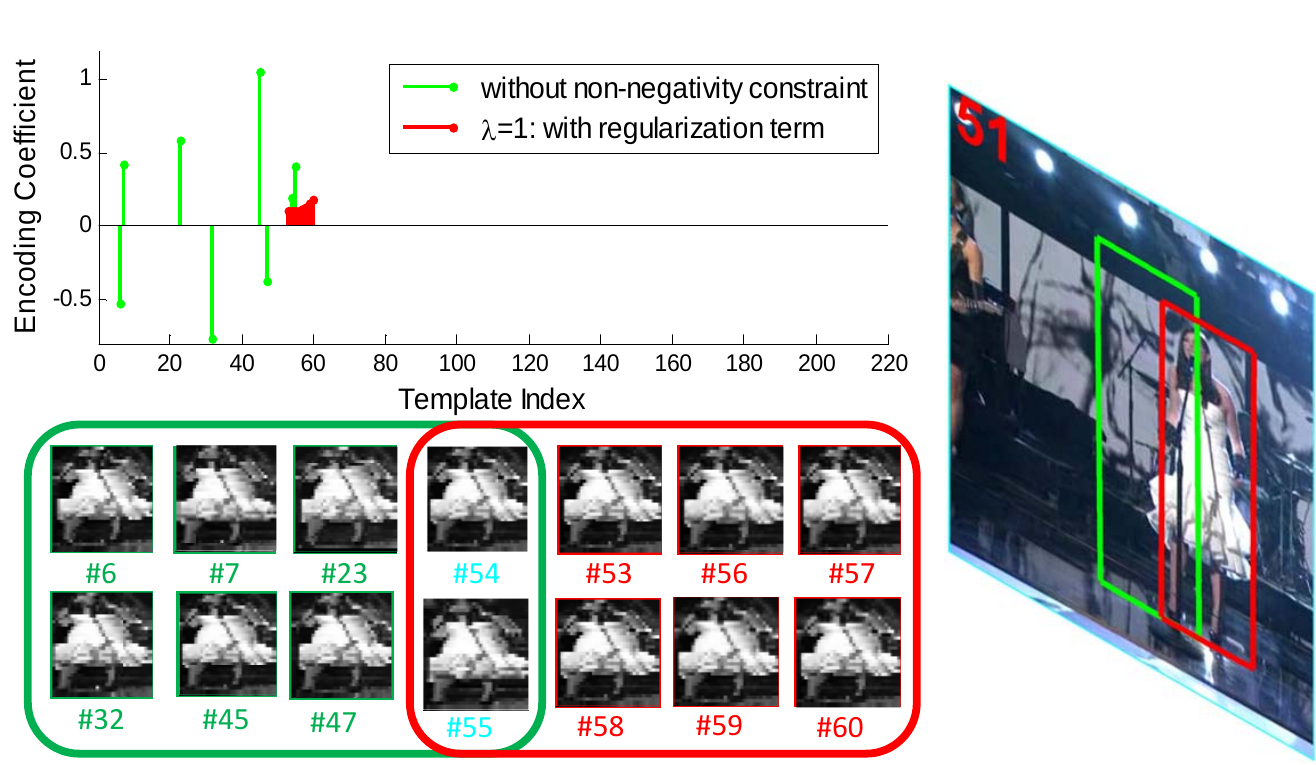}
\caption{\footnotesize Illustration of tracking result in the Sequence {\em Singer1}.
The green curves means without non-negativity constraint on encoding coefficients, leading to a bad tracking result represented by eight templates (\#6, \#7, \#23, \#32, \#45, \#47, \#52 and \#53) in the left-hand.
And an added $\ell_2$ norm regularization term leads to the good result (in red) by eight templates (\#53, \#54, \#55, \#56, \#57, \#58, \#59 and \#60) in the right-hand. }
\label{fig1}
\end{center}\vspace{-0.8cm}
\end{figure}
The experiment in Fig.\ref{fig1} shows the two tracking results reconstructed by different positive templates with encoding coefficients values in approximated LLC, where Template Index from \#1 to \#60 represents the positive template set, and the negative template set is from \#61 to \#210.
As shown in Fig.\ref{fig1}, the green tracking box denotes a bad tracking result without non-negativity constraint.
It is represented by by a linear combination of eight templates (\#6, \#7, \#23, \#32, \#45, \#47, \#52 and \#53) with their corresponding coefficient vectors in green curves.
The red one, as a good tracking result, is indicated by eight positive templates from \#53 to \#60 in red.
Although the good or bad tracking results are both dictated by the positive template set, the bad result suffers severe drifts and contains much background information.
In this case, the only difference between these two tracking results is that some selected positive templates (\#6, \#32 and \#47) are with negative coefficients.
In other words, positive templates with negative coefficients lead to an unreliable tracking result.
It seems to be a little specious and negative coefficients lose the significance of data representation on the target.

Just as the nonnegative matrix factorization(NMF) \cite{lee2001algorithms,Wu2014}, the target appearance is modeled as nonnegative linear combinations of a set of nonnegative bases that implicitly captures structure information of the target.
Thus, for every candidate $\textbf{y}_i$, its $K$ nearest neighbors from the template set $\textbf{T}=[\textbf{T}^{pos},\textbf{T}^{neg}]$ is obtained to construct the local dictionary $\textbf{B}_i$.
The corresponding encoding vector is obtained by:\vspace{-0.1cm}
\begin{equation}
\begin{split}
&\mathop{\mathrm{min}}\limits_{{\rm \textbf{c}}_i} \|\textbf{y}_i-\textbf{B}_i\textbf{c}_i\|^2
~~s.t. ~~\textbf{c}_i \geq 0, ~~~\textbf{1}^ \mathrm{T}\textbf{c}_i=1,\forall i
\end{split}
\label{allcn}
\end{equation}

\subsubsection{Various KNN for Multiple Local Dictionaries}
In most case, the number of nearest neighbors $K$ is specified manually with a fixed value.
As such, this parameter is not automatically adjusted and only one local dictionary $\textbf{B}_i$ corresponding to $\textbf{y}_i$ is selected.
To sufficiently capitalize on the intrinsic information of each $\textbf{y}_i$ and templates set $\textbf{T}$,
we can produce various local dictionaries with different numbers of nearest neighbors.
Their corresponding coefficient vectors $\textbf{c}^j_i$ are obtained by:
\begin{small}
\begin{equation}\label{knn}
\left\{
\begin{array}{rcl}
\begin{split}
&\mathop{\mathrm{min}}\limits_{{\rm \textbf{c}}^1_i} \|\textbf{y}_i-\textbf{B}^1_i\textbf{c}^1_i\|^2   &      & s.t. ~~\textbf{c}^1_i \geq 0, ~~~\textbf{1}^ \mathrm{T}\textbf{c}^1_i=1,\forall i; \\
&\mathop{\mathrm{min}}\limits_{{\rm \textbf{c}}^2_i} \|\textbf{y}_i-\textbf{B}^2_i\textbf{c}^2_i\|^2   &      & s.t. ~~\textbf{c}^2_i \geq 0, ~~~\textbf{1}^ \mathrm{T}\textbf{c}^2_i=1,\forall i;\\
&~~......\\
&\mathop{\mathrm{min}}\limits_{{\rm \textbf{c}}^j_i} \|\textbf{y}_i-\textbf{B}^j_i\textbf{c}^j_i\|^2   &      & s.t. ~~\textbf{c}^j_i \geq 0, ~~~\textbf{1}^ \mathrm{T}\textbf{c}^j_i=1,\forall i;\\
&~~......\\
&\mathop{\mathrm{min}}\limits_{{\rm \textbf{c}}^m_i} \|\textbf{y}_i-\textbf{B}^m_i\textbf{c}^m_i\|^2   &      & s.t. ~~\textbf{c}^m_i \geq 0, ~~~\textbf{1}^ \mathrm{T}\textbf{c}^m_i=1,\forall i.\\
\end{split}
\end{array} \right.
\end{equation}
\end{small}
\noindent where $m$ is the number of various dictionaries ($\textbf{B}^1_i, \textbf{B}^2_i, ..., \textbf{B}^m_i$).
And $\textbf{B}^j_i \in \mathbb{R}^{M\times k_j}$ is constructed by the candidate $\textbf{y}_i$'s $k_j$ nearest neighbors from the templates set $\textbf{T}$.
After obtaining their corresponding encoding vectors $\textbf{c}^j_i$, how to combine these $\textbf{c}^j_i$ to the final encoding vector $\textbf{c}_i$ is an overriding concern in our tracking framework.
Let $\textbf{w}=[w_1,w_2,...,w_m]^T$ be a weight vector, it implies $w_1+w_2+...+w_m=1$ and $w_j\geq 0$.
The weight vector $\textbf{w}$ is obtained by the following objective function:
\begin{equation}
\begin{split}
&\mathop{\mathrm{min}}\limits_{\textbf{w}} \|\textbf{y}_i- \sum_{j=1}^{m} w_j(\textbf{B}_i^j\textbf{c}^j_i)\|^2 \\
&~s.t. ~~\textbf{1}^ \mathrm{T}\textbf{w}=1,~~~w_j \geq 0 ~~~~\forall j \\
\end{split}
\label{multi}
\end{equation}
Let $\textbf{D}=[\textbf{B}_i^1\textbf{c}^1_i, \textbf{B}_i^2\textbf{c}^2_i,...,\textbf{B}_i^m\textbf{c}^m_i]$,
(\ref{multi}) is  transformed into:
\begin{equation}
\begin{split}
&\mathop{\mathrm{min}}\limits_{\textbf{w}} \|\textbf{y}_i- \textbf{D}\textbf{w}\|^2
~~s.t.~~\textbf{1}^ \mathrm{T}\textbf{w}=1,~w_j \geq 0 ~~~~\forall j\\
\end{split}
\label{multillc}
\end{equation}
This objective function is also an approximate LLC problem with  non-negativity constraint, similar to (\ref{allcn}) and (\ref{knn}).

\subsection{Solving for these Objective Functions}
\subsubsection{Substitute Non-negativity constraint to $\ell_2$ norm}
In the above analysis, $\textbf{c}_i$ and $\textbf{w}$ are both obtained by solving the approximated LLC problem with non-negativity constraint.
There are many optimal iteration methods for the constraint linear quadratic programming problem, such as interior point method, the accelerated proximal gradient method (APG) \cite{Bao2012}, ADMM \cite{parikh2013proximal}.
However, the non-negativity constraint destroys the structure of analytic solution in approximated LLC due to its non-differentiable character.

To tackle this problem , we introduce a $\ell_2$ norm regularization term to replace the non-negativity constraint.
Therefore, (\ref{allcn}) and (\ref{multillc}) are rewritten as:
\begin{equation}
\begin{split}
&\mathop{\mathrm{min}}\limits_{{\rm \textbf{c}}_i} \|\textbf{y}_i-\textbf{B}_i\textbf{c}_i\|^2 + \lambda \|\textbf{c}_i\|^2
~~s.t. ~\textbf{1}^ \mathrm{T}\textbf{c}_i=1,\forall i
\end{split}
\label{allcr}
\end{equation}
\vspace{-0.5cm}
\begin{equation}
\begin{split}
&\mathop{\mathrm{min}}\limits_{\rm \textbf{w}} \|\textbf{y}_i- \textbf{D}\textbf{w}\|^2+ \beta \|\textbf{w}\|^2
~~s.t.~\textbf{1}^ \mathrm{T}\textbf{w}=1, \forall j\\
\end{split}
\label{allcw}
\end{equation}
The solution of the objective function in (\ref{allcr}) is $\textbf{c}_i = [(\textbf{B}^\mathrm{T}_i -  \textbf{1}\textbf{y}^\mathrm{T}_i)(\textbf{B}_i-\textbf{y}_i\textbf{1}^\mathrm{T})+\lambda \textbf{I}]^{-1}\textbf{1}$.
And the solution of $\textbf{w}$ is in the similar fashion.
In this case, the structure of closed-form in coefficient vector $\textbf{c}_i$ and weight vector $\textbf{w}$ is preserved.
We will illustrate these elements in $\textbf{c}_i$ and $\textbf{w}$ still remain nonnegative in the following sections.
The rationale for this replacement is described in more details.\vspace{-0.25cm}

\subsubsection{Theoretical Analysis of this Replacement}
For the sake of mathematical convenience and easy to use in the subsequent description, we construct a local dictionary $\textbf{B}_i$ selected from $\textbf{y}_i$'s $K$ nearest neighbors (temporary not consider multiple dictionaries $\textbf{B}_i^j$ with different nearest neighbors).

Define $\textbf{A}=(\textbf{B}_i -  \textbf{y}_i\textbf{1}^\mathrm{T})$, $\textbf{F}= (\textbf{A}^\mathrm{T}\textbf{A}+\lambda \textbf{I})$,
the solution of (\ref{allcr}) is rewritten as $\textbf{c}_i = \textbf{F}^{-1}\textbf{1}$, where $\textbf{F} \in \mathbb{R}^{K\times K}$ (and $\textbf{F}^{-1}$) is a positive definite matrix.
We denote it as $\textbf{F}\succ 0$. \vspace{-0.25cm}
\newtheorem*{theorem}{Theorem 1}
\begin{theorem}
if $\textbf{F}^{-1}\succ0$ and $\textbf{F}^{-1}$ is a strictly diagonally dominant matrix
\footnote{$\textbf{A}=[a_{ij}]$, if $|a_{ii}|> \sum_{j=1,j\neq i}^{n}|a_{ij}|$, then \textbf{A} is called as a strictly diagonally dominant matrix.},
$\textbf{c}_i =\textbf{F}^{-1}\textbf{1}$ is nonnegative.
\end{theorem}\vspace{-0.1cm}
\noindent \emph{Proof.} Because $\textbf{F}^{-1}$ is a positive definite matrix,
elements in the dominant diagonal of $\textbf{F}^{-1}$ are all with positive values ($(\textbf{F}^{-1})_{ii} > 0,~~i=1,2,...,K$).

On the other hand, considering that $\textbf{F}^{-1}$ is a strictly diagonally dominant matrix, for each row of $\textbf{F}^{-1}$, satisfies:\vspace{-0.3cm}
\begin{equation}\label{proof}
|(\textbf{F}^{-1})_{jj}| > \sum_{i=1,i\neq j}^{K}|(\textbf{F}^{-1})_{ji}|~~~\forall j
\end{equation}
Noticing the above equation and we have
\begin{equation}\label{a}
\begin{split}
&\textbf{F}^{-1}\textbf{1}=\sum_{i=1}^{K}(\textbf{F}^{-1})_{ji}=\sum_{i=1,i\neq j}^{K}(\textbf{F}^{-1})_{ji}+(\textbf{F}^{-1})_{jj} \\
&~~~~~~~~> \sum_{i=1,i\neq j}^{K}(\textbf{F}^{-1})_{ji}+\sum_{i=1,i\neq j}^{K}|(\textbf{F}^{-1})_{ji}| \\
&~~~~~~~~\geq 0\\
\end{split}
\end{equation}
Thus $\textbf{c}_i =\textbf{F}^{-1}\textbf{1}>0$.
\vspace{0.2cm}

In the following, {\em the proposition that $\textbf{c}_i$ is nonnegative is converted to how to guarantee $\textbf{F}^{-1}$ as a strictly diagonally dominant matrix.}
An intuitive idea is to make $\textbf{F}$ a strictly diagonally dominant matrix by choosing $\lambda$ value ($\textbf{F}=\textbf{A}^\mathrm{T}\textbf{A}+\lambda \textbf{I}$).
And then seek for the relationship between $\textbf{F}$ and $\textbf{F}^{-1}$ in terms of strictly diagonally dominant character.
It is easy to choose $\lambda$, and the lower bound of $\lambda$ is given:
\begin{small}
\begin{equation}\label{lambda}
 \lambda \geq \max\{\sum_{j\neq 1}^{K}|(\textbf{A}^T\textbf{A})_{1j}|,\sum_{j\neq 2}^{K}|(\textbf{A}^T\textbf{A})_{2j}|,...,\sum_{j\neq K}^{K}|(\textbf{A}^T\textbf{A})_{Kj}|\}+\epsilon
\end{equation}
\end{small}
where $\epsilon$ is an arbitrarily small positive constant.
Because $\textbf{A}^\mathrm{T}\textbf{A}$ is a positive semi-definite matrix, then $\textbf{F}\in R^{K \times K}$ is not only a positive definite matrix but also strictly diagonally dominant matrix by choosing $\lambda$ to satisfy (\ref{lambda}).

In this condition, {\em the proposition that $\textbf{c}_i$ is nonnegative is transferred to prove a conclusion that  $\textbf{F}^{-1}$ is a strictly diagonally dominant matrix if and only if $\textbf{F}$ is a strictly diagonally dominant matrix.}

However, $\textbf{F}\succ 0$ and its strictly diagonally dominant property can not guarantee $\textbf{F}^{-1}$ is a diagonally dominant matrix in theory.
This proposition is tenable only when $K\leq2$, and a counterexample ($K=3$ means the dimension of $\textbf{F}$ is $3$) is constructed as shown in below:

\begin{small}
\begin{equation}
\textbf{F}^{-1}=\left(                 
  \begin{array}{ccc}   
    5 & 2 & -2\\  
    2 & 5 & 2\\  
    -2 & 2 & 5\\
  \end{array}
\right)                 
^{-1}=
\left(                 
  \begin{array}{ccc}   
    0.43 & -0.29 & 0.29\\  
    -0.29 & 0.43 & -0.29\\  
    0.29 & -0.29 & 0.43\\
  \end{array}
\right)
\end{equation}
\end{small}
In the above example, it is clear that $\textbf{F}^{-1}$ is not a strictly diagonally dominant matrix despite that $\textbf{F}$ is.
\subsubsection{Rationality of this Replacement in Image Data}
In the above theoretical analysis, it seems to be unreasonable when the non-negativity constraint is replaced by $\ell_2$ norm regularization term on $\textbf{c}_i$.
However, in our practical application, some implicit information in $\textbf{B}_i$, $\textbf{A}$, and $\textbf{F}$ is overlooked.
For example, these elements in these matrices are nonnegative; $\textbf{B}_i$, $\textbf{y}_i$ have been normalized, whose  values are from $0$ to $1$.

First, we need to analyse the property of $\textbf{F}$ and $\textbf{F}^{-1}$ with respect to a growth tendency to $\lambda$.
Ostroski $et~al.$ discusses the upper bound of these elements in a strictly diagonally dominant matrix's inverse matrix \cite{ostrowski1952note}.
Let $\textbf{F}=[f]_{ij}$, and
\begin{equation}\label{rate}
  \mu_i=\frac{1}{|f_{ii}|} \sum_{i=1,j\neq i}^K |f_{ij}|,~~0\leq \mu_i < 1,~~i=1,2,...,K.
\end{equation}
Then elements in the dominant diagonal of $\textbf{F}^{-1}$ satisfies:
\begin{equation}\label{range}
  \frac{1}{|f_{jj}|(1+\mu_j)}\leq (\textbf{F}^{-1})_{jj} \leq \frac{1}{|f_{jj}|(1-\mu_j)}
\end{equation}

As with $\lambda$ increases, $\mu_i$ tends to decrease.
And the value range of $(\textbf{F}^{-1})_{jj}$ will reduce.
When $\lambda$ tends to a sufficient constant, $\sum_{i=1,j\neq i}^K |f_{ij}|$ pales in importance compared with $|f_{ii}|$ (means $\mu_i \rightarrow 0$).
$(\textbf{F}^{-1})_{jj}$ approximates to $1/|f_{jj}|$.
$\textbf{F}^{-1}$ is a diagonal matrix approximately.
In this case, as a diagonal matrix, $\textbf{F}^{-1}$ is definitely a diagonally dominant matrix.
In a word, by choosing an appropriate $\lambda$, $\textbf{F}^{-1}$ approximates to a diagonally dominant matrix to ensure $\textbf{c}_i$ nonnegative.

Now we should be confronted with how to select the proper $\lambda$.
That $\lambda$ is selected to a extremely large value (e.g.,$10^4$) would make no sense to our practice application.
The relatively proper $\lambda$ is mainly determined by our image data.

Considering each column of $\textbf{B}_i$ and $\textbf{y}_i  \textbf{1}^ \mathrm{T}$ ($\textbf{y}_i\in \mathbb{R}^{1024}$ represents a $32\times 32$ image patch), elements in these matrices are quite small.
\begin{figure}
\begin{center}
\includegraphics[width=0.48\textwidth]{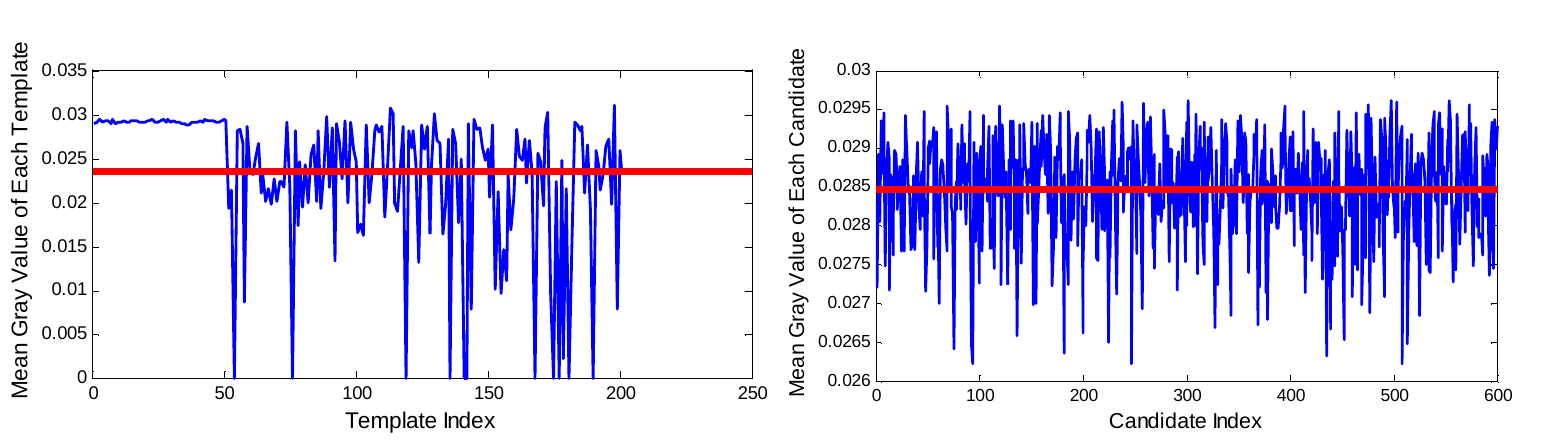}
\caption{\footnotesize Illustration of the mean gray value of each template in the left-hand figure and candidate in the right-hand figure. The red line represents the mean gray value of all templates and candidates.}
\label{fig2}
\end{center}\vspace{-0.8cm}
\end{figure}
Empirical statistic experiments\footnote{We analyse 21 sequences in VTB, and datasets available at http://visualtracking.net} in Fig.\ref{fig2} show that the mean gray value of each positive template is roughly $0.029$, and those of negative templates seem to be relatively chaotic, from $0$ to $0.3$.
Thus the mean gray value of $\textbf{B}_i$ selected from positive templates $\textbf{T}^{pos}$ is approximately equal to $0.029$.
For candidates as shown in the right-hand of Fig.\ref{fig2}, their values range from $0.026$ to $0.03$.
In this condition, each element $a_{ij}$ in $\textbf{A}=\textbf{B}_i - \textbf{y}_i\textbf{1}^T \in (-0.001,0.003)$.
The upper bound of each element in $\textbf{A}^T\textbf{A}$ is estimated to $0.003\times 0.003\times 1024\approx 0.01$.

After the upper bound of elements in $\textbf{A}^T\textbf{A}$ are obtained, $\lambda$ is easily solved.\vspace{-0.3cm}
\begin{equation}\label{est_lambda}
 \sum_{i=1,i\neq j}^{K}|f_{ij}|\leq 0.01*(K-1) \leq \lambda~~~ \forall j
\end{equation}

In sum, the lower bound of the proper $\lambda$ is obtained in (\ref{est_lambda}).
The lower bound of the proper $\beta$ is in a similar fashion.
Specially, with respect to different nearest neighbors $k_1=5,k_2=8,k_3=10$ in our experiment, $\lambda = 1$ is chosen.
This value not only satisfies (\ref{est_lambda}) but also is much larger than $0.01$.
In this condition, $\textbf{F}$ can be approximated to a diagonal matrix, accordingly, $\textbf{F}^{-1}$ can be regarded as a strictly diagonal dominant matrix.
Moreover, because the \textbf{Theorem 1} is a sufficient and unnecessary condition, relatively smaller $\lambda$ could also guarantee these elements in $\textbf{c}_i$ non-negative.
We will analyze the tracking results with different $\lambda$ values in Section \ref{sec:conclusion}.

\subsection{Encoding Vectors and Confidence Measure}
From the above analysis, it is reasonable to substitute the non-negativity constraint to $\ell_2$ norm regularization term in our image data.
Therefore $\textbf{c}_i^j \in \mathbb{R}^{k_j}$ is obtained by (\ref{allcr}) with its corresponding local dictionaries $\textbf{B}_i^j \in \mathbb{R}^{M\times k_j}$.
Noting that the selected templates composed of $\textbf{B}_i^j$ need to be recorded.
The location information of these templates is denoted as an indicator vector $\textbf{U}=[u_1, u_2,...,u_{k_j}]$.
It means that the first adopted template is in the $u_1$-th of the whole templates set, and until the $k_j$ adopted template is in the $u_{k_j}$-th of templates set.

Subsequently, let $\textbf{d}_i^j \in \mathbb{R}^{p+n}$ be the uniform coefficient vector, whose dimension is equal to the number of templates.
Then elements in $\textbf{c}_i^j$ are assigned to $\textbf{d}_i^j$.
The allocation rule by the indicator vector is following:
\begin{small}
\begin{equation}\label{allocation}
 \textbf{d}_i^j(u_1) := \textbf{c}_i^j(1);~~\textbf{d}_i^j(u_2) := \textbf{c}_i^j(2);......~~\textbf{d}_i^j(u_{k_j}) := \textbf{c}_i^j(k_j)
\end{equation}
\end{small}
where the remaining elements in $\textbf{d}_i^j$ are filled with zero.

After weight vector $\textbf{w}$ and uniform coefficient vector $\textbf{d}_i^j$ are obtained, the output encoding vector $\textbf{d}_i = \textbf{w}^T\textbf{d}_i^j$.
The algorithm for solving the output encoding vector $\textbf{d}_i$ is described in {\bf Algorithm 1}.

\begin{tabular}[h]{p{0.9\columnwidth}}
  \toprule
  \noindent \textbf{Algorithm 1.} Algorithm for solving $\textbf{d}_i$. \\
  \midrule
  \footnotesize
  \footnotesize \textbf{Input:} the candidate $\textbf{y}_i$, dictionaries  with different nearest neighbors: $\textbf{B}_i^1$, $\textbf{B}_i^2$,...,$\textbf{B}_i^m$,  iteration times T \\
  \footnotesize \textbf{Output:} the output encoding coefficient $\textbf{d}_i$\\
  \hline
  \footnotesize 1. \textbf{Initialization:} $w_1=w_2=...=w_m=1/m$ \\
  \footnotesize 2. \textbf{for} $ j=1,2,...,m$  \textbf{do}\\
  \footnotesize  ~~~~2.1 $\textbf{c}_i^j$ is obtained by (\ref{allcr}).\\
  \footnotesize  ~~~~2.2. the uniform coefficient vector $\textbf{d}_i^j$ is obtained by (\ref{allocation}). \\
  \footnotesize 2. \textbf{end for} \\
  \footnotesize 3. \textbf{for} $k=1,2,...,T$  \textbf{do}\\
  \footnotesize ~~~~~3.1. by fixing $\textbf{w}$, $\textbf{d}_i  =  \sum_{j=1}^{m}w_j\textbf{d}_i^j$ \\
  \footnotesize ~~~~~3.2. by fixing $\textbf{d}_i$, $\textbf{w}$ is solved by Eq.(\ref{allcw}). \\
  \footnotesize 3. \textbf{end  for}\\
  \bottomrule
  \label{flow}
\end{tabular}\vspace{-0.3cm}

$\textbf{d}_i$ is divided into two parts: $\textbf{d}_i=[\textbf{d}_i^{pos},\textbf{d}_i^{neg}]$ with respect to $\textbf{T}^{pos}$ and $\textbf{T}^{neg}$.
Thus, we formulate the confidence value $h_i$ of its corresponding candidate $\textbf{y}_i$:
\begin{equation}\label{conf}
h_i = \frac{1}{exp(-\alpha(\varepsilon^{pos}_i - \varepsilon^{neg}_i))}
\end{equation}
where $\varepsilon^{pos}_i=\|\textbf{y}_i-\textbf{T}^{pos} \textbf{d}^{pos}_i\|^2$ is the reconstruction error of the candidate $\textbf{y}_i$ with the positive template set, and $\textbf{d}^{pos}_i$ is corresponding coefficients.
Similarly, $\varepsilon^{neg}_i = \|\textbf{y}_i-\textbf{T}^{neg} \textbf{d}^{neg}_i\|^2$ is the reconstruction error of the candidate $\textbf{y}_i$ with the negative template set,
and $\textbf{d}^{neg}_i$ is related coefficients.
The parameter $\alpha$ is a normalization factor, fixed to 2.5 in our experiments.
The optimal state $x^{*}_t$ of frame $t$ is the candidate with the highest probability in $\textbf{H}=[h_1,h_2,...,h_N]$.\\

\subsection{Occlusion Detection and Model Update}
\label{sec:typestyle}
\begin{figure}
\begin{center}
\includegraphics[width=0.48\textwidth]{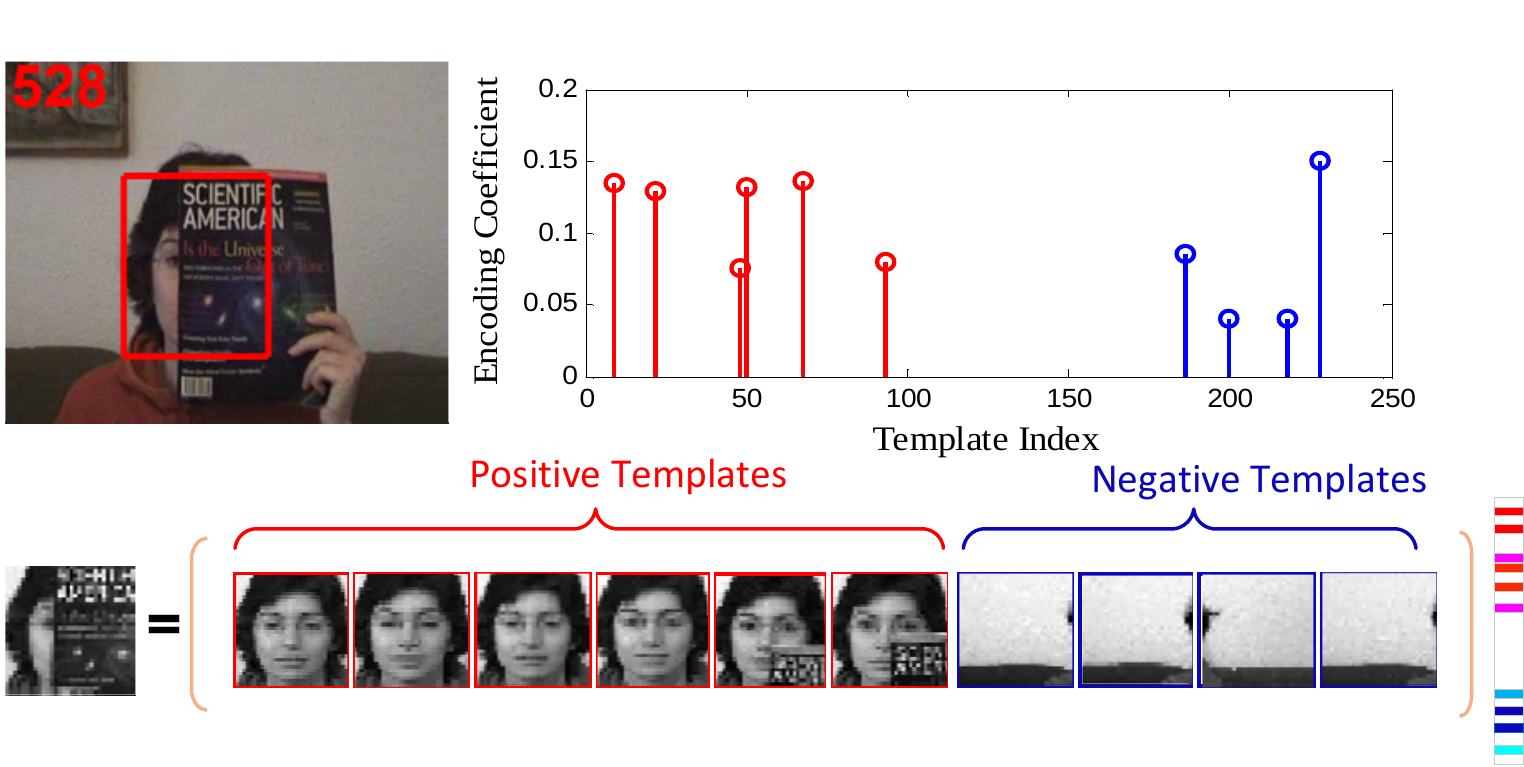}
\caption{\footnotesize Illustration of an occluded target is represented by positive templates and negative templates.}
\label{occ}
\end{center}\vspace{-0.8cm}
\end{figure}
In tracking processing, under the condition of no or slight occlusion, the target should be entirely represented by positive templates.
If the object suffers severe occlusion, it is reconstructed by not only the positive template set but also the negative template set.

Based on this, we propose a heuristic method to detect relatively larger occlusion.
The occlusion detection criterion is mainly involved with whether negative templates are used to represent the target.
If several negative templates are used to reconstruct the target, it means that the target suffers from severe occlusions in high probability, and vice versa.
The target is regarded as suffering from severe occlusions when more than one negative template are utilized to reconstruct the target, just to decrease error detection rate.
For the sake of mathematical convenience, we denote the number of negative templates representing the target as $LEN(neg_*)$.

The result of the occluded target reconstructed by template sets is shown in Fig.\ref{occ}.
For example, at \#528th frame, the severely occluded target is represented by six positive templates and four negative templates.
Four negative templates are used to represent the target, which means that the target is heavily occluded.
Therefore, the experimental results verify the validity of occlusion detection criterion.

Normally when an occlusion is detected, the positive template set should not be updated while the negative template set is updated regularly every 5 frames.
If the reconstruction error with the positive template set $\varepsilon^{pos}_*$ is smaller than a pre-defined threshold, such as 0.1, we conclude that the current tracking result is a good candidate to represent the target in the following sequence.
Thus the good tracking result is added into the positive template set.
Along with these good tracking results continuously added in, the size of positive template set becomes larger.
To avoid higher computational complexity, a template in positive template set closest to the newly added good tracking result is substituted when the number of positive template reaches 100.

The proposed tracker use approximated LLC for finding encoding vectors based on particle filter framework. The flowchart of the tracking algorithm is summarized in   \textbf{Algorithm 2}.

\vspace{0.1cm}
\begin{tabular}{p{0.9\columnwidth}}
\toprule
  \noindent \footnotesize \textbf{Algorithm 2.} Algorithm for Our proposed Tracker. \\
  \midrule
 \footnotesize 1. Initialization: Extract templates $\textbf{T}^{pos}$, $\textbf{T}^{neg}$ in the 1st frame.\\
 \footnotesize 2. \textbf{for} t = 2 to the end of the sequence \\
 \footnotesize ~~~~2.1. $N$ particles $\textbf{Y}_{1:N}$ are sampled. \\
 \footnotesize ~~~~2.2. Construct different dictionaries with different \\
 \footnotesize ~~~~~~~~~~~~($k_1,k_2,...,k_m$) nearest neighbors: $\textbf{B}_i^1$, $\textbf{B}_i^2$,...,$\textbf{B}_i^m$.\\
 \footnotesize ~~~~2.3. \textbf{for} $i=1:N$ \\
 \footnotesize ~~~~~~~~2.3.1 Solve the output encoding vector $\textbf{d}_i$ by $\textbf{Algorithm 1}$.\\
 \footnotesize ~~~~~~~~2.3.2 Calculate confidence value $h_i$ of $\textbf{y}_i$ by (\ref{conf}). \\
 \footnotesize ~~~~2.3. \textbf{end for} \\
 \footnotesize ~~~~2.4. Chose the optimal state $\textbf{x}^*_t$ by the highest confidence value. \\
 \footnotesize ~~~~2.5. Update: \textbf{for} every 5 frames \\
 \footnotesize ~~~~~~~~~ 2.5.1 Update negative templates \\
 \footnotesize ~~~~~~~~~ 2.5.2  \textbf{if} $LEN(neg_*) \leq 2$ \\
 \footnotesize ~~~~~~~~~~~~~ 2.5.2.1 incremental update the positive template set. \\
 \footnotesize ~~~~~~~~~ 2.5.2 \textbf{end if} \\
 \footnotesize ~~~~2.5. \textbf{end for} \\
 \footnotesize  2. \textbf{end for} \vspace{-0.1cm} \\
  \bottomrule
\end{tabular}

\subsection{Further Analysis on Regularization term}
 Based on the above analysis, non-negative property of $\textbf{c}_i$ are involved with choosing an appropriate $\lambda$.

From machine learning view, the objective function in (\ref{allcr}) can be viewed as a loss function $\|\textbf{y}_i-\textbf{B}_i\textbf{c}_i\|^2$ and its regularization term $\lambda \|\textbf{c}_i\|^2$.
Only considering the regularization term and its shift-invariant constraint:\vspace{-0.3cm}
 \begin{equation}
\mathop{\mathrm{min}}~~\lambda \|\textbf{c}_i\|^2
~~~s.t. \textbf{1}^ \mathrm{T}\textbf{c}_i=1,\forall i
\label{meine}
\end{equation}
This is an average inequality and the minimum is following:
 \begin{equation}\vspace{-0.6cm}
  \begin{split}
&\textbf{c}^\mathrm{T}_i \textbf{c}_i  = \sum_{j=1}^{K}c_{i(j)}^2 = c_{i(1)}^2+c_{i(2)}^2+...+c_{i(K)}^2   \\
&~~~~~\geq (\frac {c_{i(1)}+c_{i(2)}+...+c_{i(K)}}{K})^2= \frac{1}{K^2} \\
\label{meavg}
  \end{split}
\end{equation}
The inequality achieves the minimal value if and only if $c_{i(1)}=c_{i(2)}=...=c_{i(K)}=\frac{1}{K}$.
It illustrates that these nonnegative elements in $\textbf{c}_i$ tend to be approximately equal by adjusting the regularization parameter $\lambda$.
The larger regularization parameter is, the more obvious average effect shown in Eq.(\ref{meavg})  on encoding coefficients have.
On the other hand, by adding this $\ell_2$ norm regularization term, the model effectively avoids over-fitting.
As with $\lambda$ increases, the bias of the model increases and the variance falls down.
The disparity and diversity of the output $\textbf{c}_i$ compared with its expectation would decrease.
In other words, $\textbf{c}_i$ is relatively more stable.

Experiments about the benefit of the additional $\ell_2$ norm regularization term $\lambda \|\textbf{c}_i\|^2$ seem clearly to be the litmus test for our discussions as shown in Fig.\ref{fig1}.
The red one with $\lambda=1$ in Fig.\ref{fig1} shows that the target is approximately equally reconstructed by the local dictionary composed of eight positive templates.
These eight positive templates takes the same and equal effect on representing the target.
We think this phenomenon that the target should be equally represented by several positive templates have distinct physical meanings.

In sum, the added $\ell_2$ norm regularization term leads to many advantages more than avoidance of over-fitting.
Especially for high-dimension data (e.g., our image data), in approximated LLC, the non-negativity constraint is entirely substituted by the $\ell_2$ norm regularization with adjusting an appropriate regularization parameter $\lambda$.
And this not only fits our computer vision applications but also will be well done in other fields.

\section{Experiments}
\label{sec:experiment}
\begin{figure*}
\begin{center}
\includegraphics[width=1.0\textwidth]{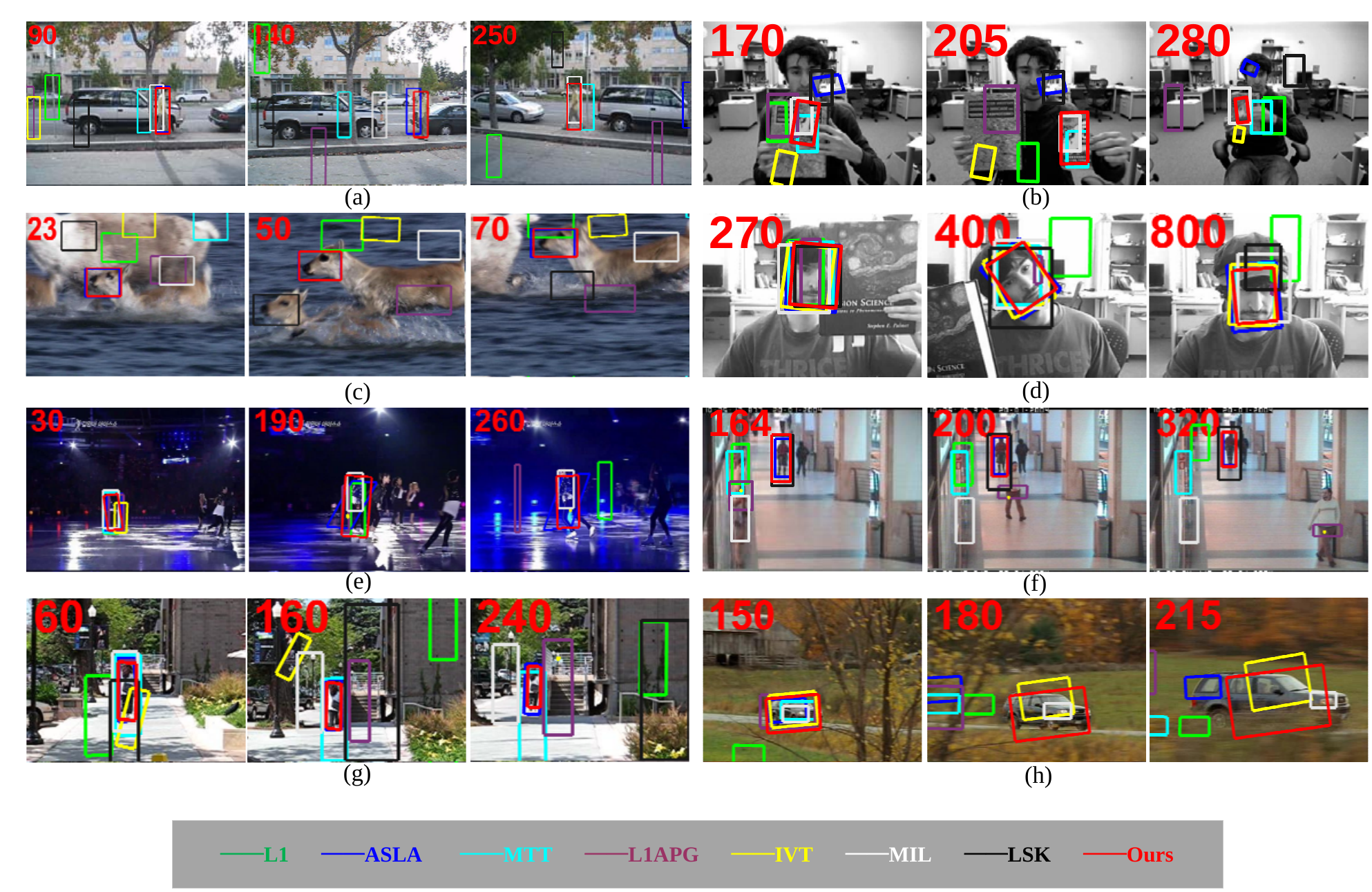}
\caption{\footnotesize Representative frames of some sampled tracking results. And subfigures from top to bottom, left to right: (a) - (h), from video DavidOutdoor,~ ClifBar,~ Deer, ~Occlusion2, ~Skating1, ~Caviar1, ~Human7 and Carscale.}
\label{result}
\end{center}\vspace{-0.3cm}
\end{figure*}

%
%

\begin{table}
\centering
\scriptsize
\caption{\footnotesize Performance in terms of "center location error" (CLE) in pixels.  \textcolor[rgb]{1,0,0}{Red} and \textcolor[rgb]{0,0,1}{blue} colors indicate the best and 2nd best performance, respectively.}
\vspace{0.1cm}
\begin{tabular}{|c|p{0.42cm}|p{0.4cm}|p{0.62cm}|p{0.4cm}|p{0.4cm}|p{0.4cm}|p{0.45cm}|p{0.32cm}|p{0.32cm}|}
\hline
  Sequence &LSK	&L1	&L1APG	&MTT	&IVT	&MIL	&ASLA &KCF &Ours \\
  \hline
  Car4 & 258.6 & 4.1 & 223.0 & 96.0 &\textcolor[rgb]{1,0,0}{2.6} & 60.1 &\textcolor[rgb]{0,0,1}{3.5} &19.1 & 4.3 \\
Carscale & 34.5 & 66.8 & 81.2 & 83.7 &\textcolor[rgb]{0,0,1}{11.7} & 27.3 & 48.8 &43.0 &\textcolor[rgb]{1,0,0}{10.9}\\
David3	 & 162.6 & 100.4 & 204.4 & 363.3 & 100.2 &38.4 & 87.4 &\textcolor[rgb]{1,0,0}{5.5} &\textcolor[rgb]{0,0,1}{6.0}\\
Deer	 & 117.0 & 97.9 & 197.9 & 15.2 & 123.8 & 225.8 &\textcolor[rgb]{1,0,0}{10.7} &23.4 &\textcolor[rgb]{0,0,1}{12.3}\\
Faceocc2
	 & 19.1 & 11.1 & 16.2 & 9.9 & 8.4 & 14.1 &\textcolor[rgb]{1,0,0}{3.5} &10.5 &\textcolor[rgb]{0,0,1}{7.5}\\
Caviar3
	 & 23.7 & 65.9 & 26.6 & 64.8 & 66.2 & 57.8 &\textcolor[rgb]{1,0,0}{2.2} &23.4 &\textcolor[rgb]{0,0,1}{17.7}\\

ClifBar
	 & 71.1 & 75.9 & 72.1 & 35.7 & 62.6 &\textcolor[rgb]{0,0,1}{7.8} & 61.1 &41.6 &\textcolor[rgb]{1,0,0}{5.0}\\

Human7
	 & 51.4 & 103.7 & 27.8 & 17.0 & 54.1 & 21.9 &\textcolor[rgb]{1,0,0}{2.9} &57.6 &\textcolor[rgb]{0,0,1}{3.4}\\

Faceocc1
	 &\textcolor[rgb]{1,0,0}{5.5} & 6.5 & 8.3 & 17.3 & 11.7 & 32.2 & 7.7 &77.3 &\textcolor[rgb]{0,0,1}{6.4}\\

Skating1
	 & 90.5 & 32.6 & 145.4 & 256.3 &31.9 & 86.2 & 47.2 &\textcolor[rgb]{0,0,1}{22.7} &\textcolor[rgb]{1,0,0}{10.7}\\

Caviar1
	 & 8.0 & 34.6 & 95.0 & 55.2 & 98.5 & 88.2 &\textcolor[rgb]{1,0,0}{1.6} &\textcolor[rgb]{0,0,1}{4.9} &4.9\\

Singer1 & 193.9 & 87.8 &\textcolor[rgb]{1,0,0}{4.6} & 16.6 & 11.4 & 15.2 &\textcolor[rgb]{0,0,1}{5.1} &14.0 & 8.3 \\
\hline
	avg. & 86.3 & 57.3 & 91.9 & 85.9 & 48.6 & 56.2 &\textcolor[rgb]{0,0,1}{23.5} &28.6 &\textcolor[rgb]{1,0,0}{8.1}\\
    \hline
\end{tabular}
\label{tabcle}
\vspace{-0.5cm}
\end{table}

\begin{table}
\centering
\scriptsize
\caption{\footnotesize Performance in terms of "overlap rate" $e$ (in pixels). \textcolor[rgb]{1,0,0}{Red} and \textcolor[rgb]{0,0,1}{blue} colors indicate the best and 2nd best performance, respectively.}
\vspace{0.1cm}
\begin{tabular}{|c|p{0.42cm}|p{0.4cm}|p{0.62cm}|p{0.4cm}|p{0.4cm}|p{0.4cm}|p{0.45cm}|p{0.32cm}|p{0.32cm}|}
\hline
  Sequence &LSK	&L1	&L1APG	&MTT	&IVT	&MIL	&ASLA &KCF &Ours \\
  \hline
  Car4     & 0.05 & 0.84 & 0.15 & 0.45 &\textcolor[rgb]{0,0,1}{0.91} & 0.34 &\textcolor[rgb]{1,0,0}{0.91} & 0.47 & 0.90 \\
	Carscale & 0.51 & 0.36 & 0.55 & 0.49 &\textcolor[rgb]{0,0,1}{0.62} & 0.42 & 0.45 &0.41 &\textcolor[rgb]{1,0,0}{0.70}\\
	David3 & 0.12 & 0.35 & 0.14 & 0.09 & 0.31 & 0.41 &0.46 &\textcolor[rgb]{0,0,1}{0.75} &\textcolor[rgb]{1,0,0}{0.75}\\
	Deer & 0.21 & 0.07 & 0.05 & 0.55 & 0.04 & 0.04 &\textcolor[rgb]{1,0,0}{0.60} &\textcolor[rgb]{0,0,1}{0.60} &0.57\\
	Faceocc2 & 0.56 & 0.67 & 0.34 & 0.70 & 0.74 & 0.61 &\textcolor[rgb]{1,0,0}{0.80} &0.72 &\textcolor[rgb]{0,0,1}{0.76}\\
	Caviar3 & 0.36 & 0.20 & 0.19 & 0.14 & 0.13 & 0.11 &\textcolor[rgb]{1,0,0}{0.84} &0.14 &\textcolor[rgb]{0,0,1}{0.48}\\
	ClifBar & 0.12 & 0.20 & 0.27 & 0.29 & 0.11 &\textcolor[rgb]{0,0,1}{0.53} & 0.21 &0.25 &\textcolor[rgb]{1,0,0}{0.65}\\
	Human7 & 0.17 & 0.05 & 0.27 & 0.48 & 0.11 & 0.29 &\textcolor[rgb]{1,0,0}{0.81} &0.28 &\textcolor[rgb]{0,0,1}{0.77}\\
	Faceocc1 & 0.82 &\textcolor[rgb]{0,0,1}{0.88} & 0.84 & 0.72 & 0.82 & 0.59 & 0.87 &0.10 &\textcolor[rgb]{1,0,0}{0.88}\\
	Skating1 & 0.28 & 0.39 & 0.10 & 0.10 & 0.06 & 0.31 & 0.40 &\textcolor[rgb]{0,0,1}{0.45} &\textcolor[rgb]{1,0,0}{0.50}\\
	Caviar1 & 0.55 & 0.28 & 0.28 & 0.28 & 0.27 & 0.25 &\textcolor[rgb]{1,0,0}{0.89} &0.69 &\textcolor[rgb]{0,0,1}{0.80}\\
	Singer1 & 0.21 & 0.24 &\textcolor[rgb]{0,0,1}{0.77} & 0.42 & 0.54 & 0.34 &\textcolor[rgb]{1,0,0}{0.79} &0.36 & 0.65 \\
\hline
	avg. & 0.33 & 0.38 & 0.33 & 0.39 & 0.39 & 0.35 &\textcolor[rgb]{0,0,1}{0.67} &0.44 &\textcolor[rgb]{1,0,0}{0.70}\\
\hline
\hline
fps.	 &6.57 & 0.28 & 4.41 & 1.02 &16.41 & 0.86 & 0.95 & 89.8 & 1.28 \\
\hline
\end{tabular}
\label{tabor}
\vspace{-0.5cm}
\end{table}

\noindent \textbf{Setup:} The proposed tracker was implemented in MATLAB with a PC with Intel Xeon E5506 CPU (2.13 GHz) with 24 GB memory.
The following parameters were used for our tests:
each observation (i.e. patch of image) was normalized to $32\times32$ pixels;
different nearest neighbors ($k_1=5,k_2=8,k_3=10$) as the three ($m=3$) local dictionaries were selected;
and the iteration times $T$ was fixed with $3$;
the $\ell_2$ norm regularization parameters were set to $\lambda=1, \beta=0.1$;
Especially, different $\lambda$ were chosen to analyze the influence of tracking results.

To evaluate the proposed method against the state-of-the-art, 8 existing methods are selected, including KCF \cite{henriques2015high}, ASLA \cite{Jia2012}, LSK \cite{Liu2013},  L1 trakcer \cite{Mei2011}, L1APG \cite{Bao2012}, MTT \cite{Zhang2013b}, IVT \cite{Ross2008a} and MIL \cite{Babenko2011a}.

\noindent {\bf Results:} Fig.\ref{result} shows screen shots of tracking results from different trackers.
Tab.\ref{tabcle} shows the performance of these methods  based on the center location error (CLE), where a small CLE value indicates more  accurate hence better tracking.
Tab.\ref{tabor} shows the performance of different methods based on the overlap rate between the tracked bounding box and the ground truth box from these methods on several videos. The overlap rate is defined as $e=\frac{area(R_T \cap R_G)}{area(R_T \cup R_G)}$, where $R_T$ and $R_G$ are the area of tracked and ground truth box, respectively.

\subsection{Qualitative Evaluation}

\textbf{Heavy Occlusion}: Fig.\ref{result}(a), (d), (f) and (h) demonstrate that the proposed method performs well in terms of position and rotation when the target undergoes severe occlusion.
 In the David3 sequence, IVT, L1APG and MTT completely fail to track at frames \#34, \#57 and \#82.
 After David passes through the tree, ASLA and our method can effectively locate the target, whereas MIL suffers a slight drifts.
 In the Faceocc2 and Caviar1 Sequence, ASLA and our method performs better than the other methods in terms of tracking accuracy.
 When comes to the target occluded by branches in the Carscale sequences at frame \#172, most methods suffers from severe drift while MIL and IVT are far from satisfaction.
 Only our proposed method shows a preferable tracking result.
 In sum, because of the non-negativity constraint, encoding coefficients are nonnegative from the selected positive templates.
 These selected positive templates cannot be regarded as negative templates as shown in Fig.\ref{fig1}.
 Tracking results verify that the non-negativity constraint and model update scheme with the severe occlusion detection method are reasonable.

 \textbf{Shape Deformation and Rotation Variation}: The trackers are easily confused if the object has changed in appearance of the target dramatically.
 Fig.\ref{result}(b), (e) and (h) illustrate the tracking results in the ClifBar, Skating1 and Carscale sequences with scale and rotation variation of the card, the skater and the car.
 In the ClifBar sequence, our proposed method and MIL perform favorably better than other methods despite that the target undergoes severe scale and rotation variation.
 The skater in Fig.\ref{result}(e) dramatically deforms her pose.
 Only our proposed method precisely tracks the skater to some extent under the condition of complicated background and low light.
 The first half of Carscale sequences, only our method could adaptive the change in size of the car.
 At frame \#215, the car undergoes not only scale variation but also heavy occlusion and fast motion, it is difficult for our method to achieve a satisfying performance.

 \textbf{Abrupt Motion and Camera Shake}:  Fig.\ref{result}(c) and (g) shows the tracking results on the Deer and Human7 sequences.
 It is difficult to predict the location of this deer and this woman accurately when they undergo an abrupt motion and camera shake.
 Most trackers lose tracking accuracy in Human7 sequence and even suffer from severe drifting in the Deer sequence.
 Our tracker and ASLA are able to distinguish the target from their surrounding background, and handles drift problem.

In summary, our test results on these videos with heavy occlusions or motion blur have shown that the proposed tracker and ASLA are effective and robust.
In terms of sequences with scale and shape variation, only our proposed tracker performs favorably better than other methods.

\subsection{The Regularization Parameter Selection}
\begin{figure}
\begin{center}
\includegraphics[width=0.45\textwidth]{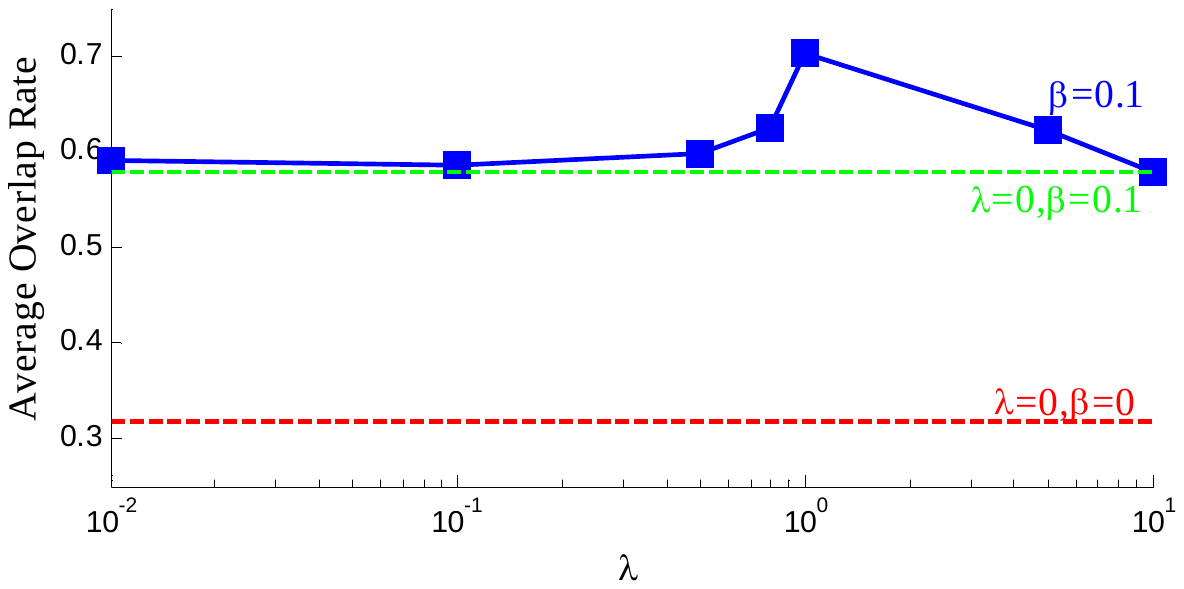}
\caption{\footnotesize The average overlap rate of our proposed tracker versus parameter $\lambda$ on these twelve sequences.}
\label{parameter}
\end{center}\vspace{-0.8cm}
\end{figure}

Fig.\ref{parameter} shows the influence of regularization parameter $\lambda$ with different values ($0,0.01,0.1,0.5,0.8,1,5,10$) on average overlap rate.
Without regularization term ($\lambda=\beta=0$) shown in the red dashed line, drifts happen in most sequences with a very low average overlap rate ($e=0.3175$), forming a sharp contrast to these methods with regularization term.
Under $\beta=0.1$, when $\lambda$ ranges from 0.01 to 10, the average overlap rate (in blue line) steadily increases and then falls down, where it reaches the summit at $\lambda=1$.
The green dashed line represents $\lambda=0,\beta=0.1$, just for a fair comparison with the blue fold line.
If the value of $\lambda$ is too small, or with image data in the same order of magnitude ($\lambda \in [0.01,0.1]$), there is slight influence on the final tracking results.
If the value of $\lambda$ is too large, it is easily over-fitting and loss function pales in importance.
It is inevitable to seek for a tradeoff between the accuracy of appearance model and the regularization term.
\section{Conclusion}
\label{sec:conclusion}

This paper proposes an effective tracker with regularization term on encoding coefficients and different numbers of nearest neighbours in the multiple local dictionaries.
The non-negativity constraint on encoding coefficients is substituted by the $\ell_2$ regularization term.
This replacement is rational for our computer vision applications.
It not only leads to these elements nonnegative, but also has an average effect on them.
These characteristics guarantee the tracking results more reliable and robust.
Moreover, the optimal convex combination of multiple local dictionaries is learned from approximated LLC.
And our occlusion detection method effectively prevents positive templates to update when the target undergoes severe occlusion.
Experimental results demonstrate that our proposed algorithm is able to track the target accurately with challenging factors.


\end{document}